%% file: acl_latex.tex
\pdfoutput=1

\documentclass[11pt]{article}

\usepackage[preprint]{acl}

\usepackage{times}
\usepackage{latexsym}

\usepackage[T1]{fontenc}

\usepackage[utf8]{inputenc}

\usepackage{microtype}

\usepackage{inconsolata}

\usepackage{graphicx}

\usepackage[utf8]{inputenc} 
\usepackage[T1]{fontenc}    
\usepackage{hyperref}       
\usepackage{url}            
\usepackage{booktabs}       
\usepackage{amsfonts}       
\usepackage{nicefrac}       
\usepackage{microtype}      
\usepackage{xcolor}         
\usepackage{multirow,multicol}
\usepackage{graphicx}
\usepackage{arydshln}
\usepackage{amsmath}
\usepackage{enumitem}

\title{\our: An IE Free Rider Hatched by Massive Nutrition in LLM's Nest}

\author{Letian Peng, Zilong Wang, Feng Yao, Jingbo Shang \\
University of California, San Diego \\
  \texttt{\{lepeng, ziw049, fengyao, jshang\}@ucsd.edu}
  }

\usepackage{xspace}

\newcommand{\our}{Cuckoo\xspace}

\begin{document}
\maketitle
\begin{abstract}
    \input{0-abs}
\end{abstract}

\input{1-intro}
\input{2-rel}
\input{3-method}
\input{4-exp}
\input{5-analysis}
\input{6-con}

\input{7-lim}

\bibliography{custom}

\clearpage

\appendix

\section{\our v.s. LLMs}
\label{apdx:vs_llm}

\begin{figure}
    \centering
    \includegraphics[width=\linewidth]{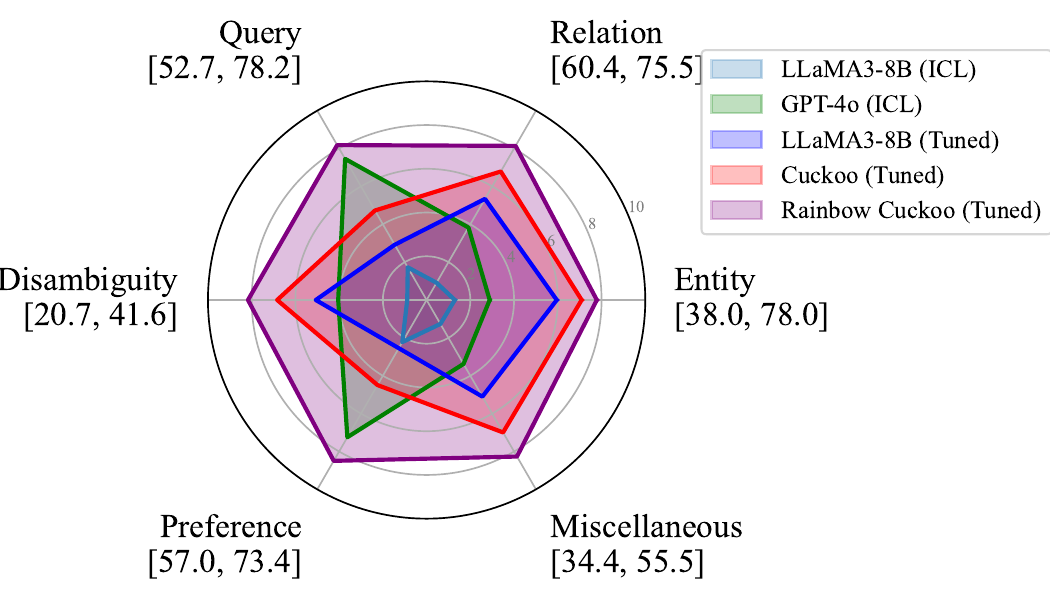}
    \caption{The performance comparison between \our and LLMs on few-shot IE performance.}
    \label{fig:llm_radar}
    \vspace{-5mm}
\end{figure}

We extend the comparison to \our versus LLMs. We select \texttt{LLaMA-3-8B-TuluV3} and \texttt{GPT-4o} to represent the fine-tunable open-source LLMs and API-based close-source LLMs. For \texttt{LLaMA-3-8B-TuluV3}, we fine-tune the LLM with the same templated data as our \our. For both LLMs, we evaluate their in-context learning IE ability based on the few shots.

We present the experiment result in Figure~\ref{fig:llm_radar}, which demonstrate that \our can outperform even fine-tuned 8B LLMs. This implicates the superior learning efficiency of NTE over NTP on IE tasks. The ICL performance of LLM significantly lags behind the fine-tuned one, restraining the performance of close-source LLMs. Finally, Rainbow \our validates itself again as the strongest few-shot IE learner even when LLMs are considered.

\paragraph{Efficiency} The time efficiency of \our is significantly higher than LLMs thanks to the specialized learning paradigm for IE. Taking NER as an example, \our is around 20$\times$ faster than \texttt{LLaMA-3-8B-TuluV3}. When the LLM is using ICL, the efficiency advantage becomes more than $50\times$, demonstrating the superior efficiency of \our. 

\section{Templates and Hyperparameters}
\label{apdx:detail}

\paragraph{Task Templates} are included in Table~\ref{tab:template}, which are used to fine-tune NTE and NTP models like \our and LLaMA on IE tasks.

\begin{table}
\centering
\small
\scalebox{0.9}{
\begin{tabular}{p{1.5cm}p{5.2cm}}
\toprule
Target & Template\\
\midrule
Entity & \textbf{User:} [Context] Question: What is the [Label] mentioned? \textbf{Assistant:} Answer: The [Label] is \\
\midrule
Relation (Kill) & \textbf{User:} [Context] Question: Who does [Entity] kill? \textbf{Assistant:} Answer: [Entity] kills \\
\midrule
Relation (Live) & \textbf{User:} [Context] Question: Where does [Entity] live in? \textbf{Assistant:} Answer: [Entity] lives in \\
\midrule
Relation (Work) & \textbf{User:} [Context] Question: Who does [Entity] work for? \textbf{Assistant:} Answer: [Entity] works for \\
\midrule
Relation (Located) & \textbf{User:} [Context] Question: Where is [Entity] located in? \textbf{Assistant:} Answer: [Entity] is located in \\
\midrule
Relation (Based) & \textbf{User:} [Context] Question: Where is [Entity] based in? \textbf{Assistant:} Answer: [Entity] is based in \\
\midrule
Relation (Adverse) & \textbf{User:} [Context] Question: What is the adverse effect of [Entity]? \textbf{Assistant:} Answer: The adverse effect of [Entity] is \\
\midrule
Query & \textbf{User:} [Context] Question: [Question] \textbf{Assistant:} Answer: \\
\midrule
Instruction (Entity) & \textbf{User:} [Context] Question: What is the [Label] mentioned? ([Instruction]) \textbf{Assistant:} Answer: The [Label] is \\
\midrule
Instruction (Query) & \textbf{User:} [Context] Question: [Question] ([Instruction]) \textbf{Assistant:} Answer: \\
\bottomrule
\end{tabular}
}
\caption{The templates used in our experiments for different tasks.} 
\vspace{-5mm}
\label{tab:template}
\end{table}

\paragraph{Hyperparameter} All models are fully fine-tuned except for \texttt{LLaMA-3-8B-TuluV3}, which exhibits a poor performance without LoRA~\citep{lora}. We use a $128$-dimension LoRA for \texttt{LLaMA-3-8B-TuluV3}. All fine-tuning uses AdamW~\citep{AdamW} as the optimizer, learning rate initialized as $1\times 10^{-5}$ to fully fine-tune RoBERTa and OPT, and $2\times 10^{-4}$ to fine-tune the LoRA. The batch size is set to $64$ for all fine-tuning. 

\section{Benchmark Details}
\label{apdx:itie}

All results in the main experiments are an average of $3$ runs on different subsets of a few shots. MRC results are evaluated on the validation split as in previous works. Instruction-following IE only focuses on the modified entity types like organization and miscellaneous.

\paragraph{Relation Extraction} gives the ground-truth entities to extract related entities. We don't run end-to-end experiments to avoid mixing entity and relation extraction abilities.

\paragraph{Duplicates} When an entity is extracted as multiple types in NER, we keep all of them because modern generative IE models (e.g., LLM) allow such features to fit into a broader usage. For instance, an LLM would say ``Kobe Bryant'' to be both a ``person'' and a ``basketball player''. For MRC, when multiple answers are extracted, we will select the answer that appears the most.

\paragraph{SQuAD-V2} is a special MRC dataset that contains unanswerable questions. We follow the initial evaluation to assign $1.0$ F1 score to abstain for these questions and $0.0$ F1 score for any answer. Adaptive training for SQuAD-V2 contains extra $32$-shot unanswerable questions.

\paragraph{Disambiguation} The $3$ instructions used for disambiguation are presented in Table~\ref{tab:instruction}. We use the follow template to prompt \texttt{GPT-4o} for filtering.

\textit{[Instruction] Does ``[Entity]'' in ``[Context]'' satisfy the definition above? Answer ``yes'' or ``no'' only.}

We manually check the filtering quality of $50$ random cases for each instruction, and find a high filtering quality of $134/150=89.33\%$.

\paragraph{Miscellaneous} For CoNLL2003, as there is already a miscellaneous type, we manually write an instruction to define the scope of miscellaneous. For MIT-Restaurant dataset, we combine ``amenity'', ``hours'', and ``price'' entity types. For MIT-Movie dataset, we combine ``actor'', ``soundtrack'', and ``quote'' entity types. Then we simply collect those types of entities to build the miscellaneous type for the benchmark. In the instruction, we include negations of miscellaneous as distractors to increase the difficulty in instruction-following.

\begin{table}
\centering
\small
\scalebox{0.9}{
\begin{tabular}{p{0.8cm}p{1.5cm}p{4cm}}
\toprule
Task & Dataset & Instruction\\
\midrule
Disamb. & CoNLL2003 & The organization entity must be a subject of any active action in the context. \\
\cmidrule(lr){2-3}
& BioBLP2004 & The provided context must contain some descriptive information about the protein. \\
\cmidrule(lr){2-3}
& Restaurant & The rating should describe a food or drink mentioned in the sentence. \\
\midrule
Prefer. & SQuAD & Give the longest answer \\
& & Give the shortest answer \\
& & Give a concise answer \\
\midrule
Misc. & CoNLL2003 & Miscellaneous includes events, nationalities and products but not person, location or organization. \\
\cmidrule(lr){2-3}
 & Restaurant & Miscellaneous includes amenity, hours and price but not rating, dish, or location. \\
\cmidrule(lr){2-3}
  & Movie & Miscellaneous includes actor, soundtrack and quote but not director, opinion, or plot. \\
\bottomrule
\end{tabular}
}
\caption{The specific instructions used for instruction-following IE.} 
\vspace{-5mm}
\label{tab:instruction}
\end{table}

The specific instructions used for instruction-following IE are listed in Table~\ref{tab:instruction}.

\section{Adaptive Supervision Scaling}
\label{apdx:adaptive_scaling}

\begin{figure}
    \centering
    \includegraphics[width=\linewidth]{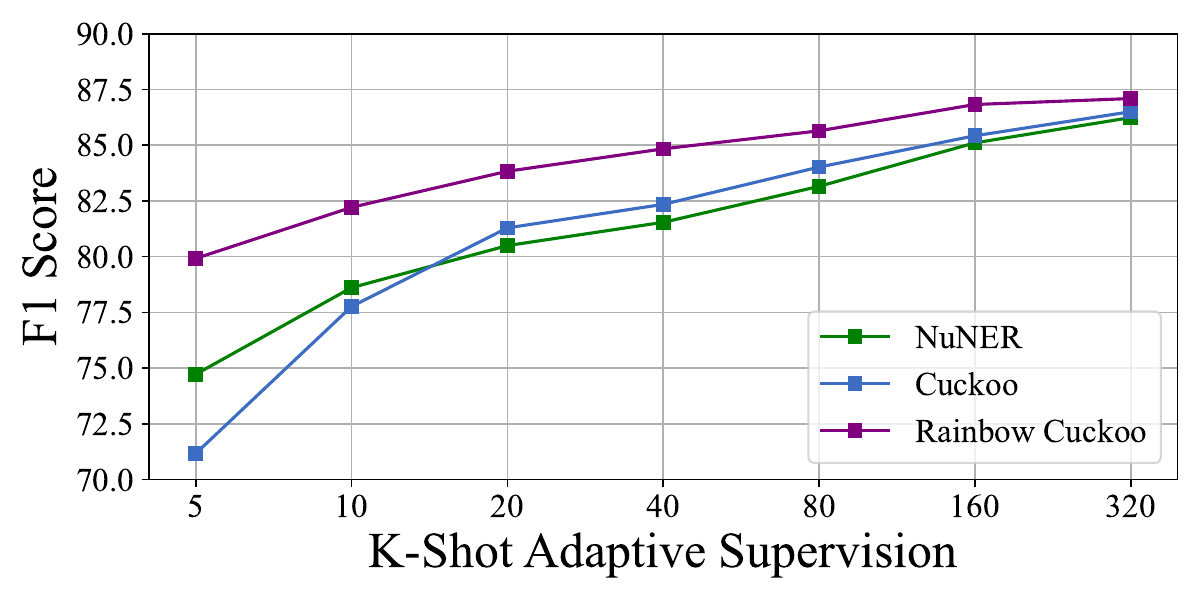}
    \vspace{-8mm}
    \caption{The scaling-up performance on adaptive supervision from CoNLL2003 of pre-trained IE models.}
    \label{fig:adaptive_scaling}
\end{figure}

In the application for IE, it's common to scale up the adaptive supervision (few-shot instances) to strengthen the model's IE ability. We plot such an example for CoNLL2003 in Figure~\ref{fig:adaptive_scaling} for transferring learning with different scales of supervision, from $5$-shot to $320$-shot. For comparison, we include the strongest NER baseline, NuNER, from the main experiment.

The results demonstrate that \our can scale up similarly as NuNER, the in-domain transfer of NuNER shows its advantage under very weak supervision but is surpassed by \our when the adaptive supervision is enough for domain understanding. Finally, Rainbow \our consistently show advantages under different adaptive supervision scales.

\section{Robustness to Verbalization}

\begin{table}
\centering
\small
\scalebox{0.9}{
\begin{tabular}{p{1.0cm}p{6.cm}}
\toprule
Rephrase & New Template/Label \\
\midrule
Template & \textbf{User:} [Context] Instruction: Extract [Label] from the text above. \textbf{Assistant:} [Label]: \\
\cmidrule(lr){2-2}
& \textbf{User:} List all [Label] entities: [Context] \textbf{Assistant:} Here are [Label] entities: 1. \\
\midrule
Label & (CoNLL2003) Person $\rightarrow$ Name\\
\cmidrule(lr){2-2}
 & (BioBLP2004) DNA $\rightarrow$ Deoxyribonucleic acid\\
\cmidrule(lr){2-2}
 & (Restaurant) Rating $\rightarrow$ Recommendation\\
\cmidrule(lr){2-2}
 & (Movie) Genre $\rightarrow$ Category\\
\bottomrule
\end{tabular}
}
\caption{The template/label variants used for robustness testing.} 
\vspace{-5mm}
\label{tab:variant}
\end{table}

As \our relies on prompts to perform different tasks. Its robustness to different verbalization of tasks and labels needs more emphasis. We select NER as an example and rephrase templates and labels in our experiments, which are listed in Table~\ref{tab:variant}. We rerun the experiments with these modifications and find the NER performance is not significantly (defined as $p < 0.05$ in significance testing) different from the initial results. This indicates \our to be robustness to different verbalization styles.

\end{document}

%% file: 0-abs.tex
Massive high-quality data, both pre-training raw texts and post-training annotations, have been carefully prepared to incubate advanced large language models (LLMs).
In contrast, for information extraction (IE), pre-training data, such as BIO-tagged sequences, are hard to scale up. 
We show that IE models can act as free riders on LLM resources by reframing next-token \emph{prediction} into \emph{extraction} for tokens already present in the context.
Specifically, our proposed next tokens extraction (NTE) paradigm learns a versatile IE model, \emph{\our\footnote{\our is known for laying its eggs in other birds’ nests, tricking them into raising its chicks.}}, with $102.6$M extractive data converted from LLM's pre-training and post-training data. 
Under the few-shot setting, \our adapts effectively to traditional and complex instruction-following IE with better performance than existing pre-trained IE models.
As a free rider, \our can naturally evolve with the ongoing advancements in LLM data preparation, benefiting from improvements in LLM training pipelines without additional manual effort.\footnote{Open \our: \href{https://github.com/KomeijiForce/Cuckoo}{https://github.com/KomeijiForce/Cuckoo}}

%% file: 1-intro.tex
\section{Introduction}

The biggest lesson researchers have learned from training large language models (LLMs)~\citep{tulu,llama-2,achiam2023gpt4,olmo,dubey2024llama3,team2024gemma} is the power of massive and high-quality data~\citep{scaling_law,scaling_law_transfer}. Although pre-training information extraction (IE) models~\citep{pretrain_ner,multinerd,UIE,TadNER,NuNER,metaie} has once been a popular topic before the rise of general LLMs, the relative scarcity of automated annotations has limited the further development of this domain. Consequently, more and more researchers have accepted LLMs as backbone models for IE tasks~\citep{llm4clinicalie,gpt-ner,llm4ie}.

The primary reason for the temporary lag in IE pre-training is the stricter format requirements for data collection compared to those for LLMs.
The paradigm for learning LLMs, the next token prediction (NTP), can utilize every token in the sentence as an annotation. In contrast, IE pre-training always requires spans annotated with label names. 
While certain platforms provide massive annotations, such as Page Links in Wikipedia~\citep{ner_wiki,fewnerd,fewrel,multinerd}, they are still much less efficient than NTP. 
To illustrate the gap, Multinerd~\citep{multinerd} takes multiple processing efforts to collect $164$K English named entity recognition (NER) instances from Wikipedia and Wikinews, while NTP can easily gather trillions of tokens from raw texts as supervision. 

\begin{figure*}
    \centering
    \includegraphics[width=\linewidth]{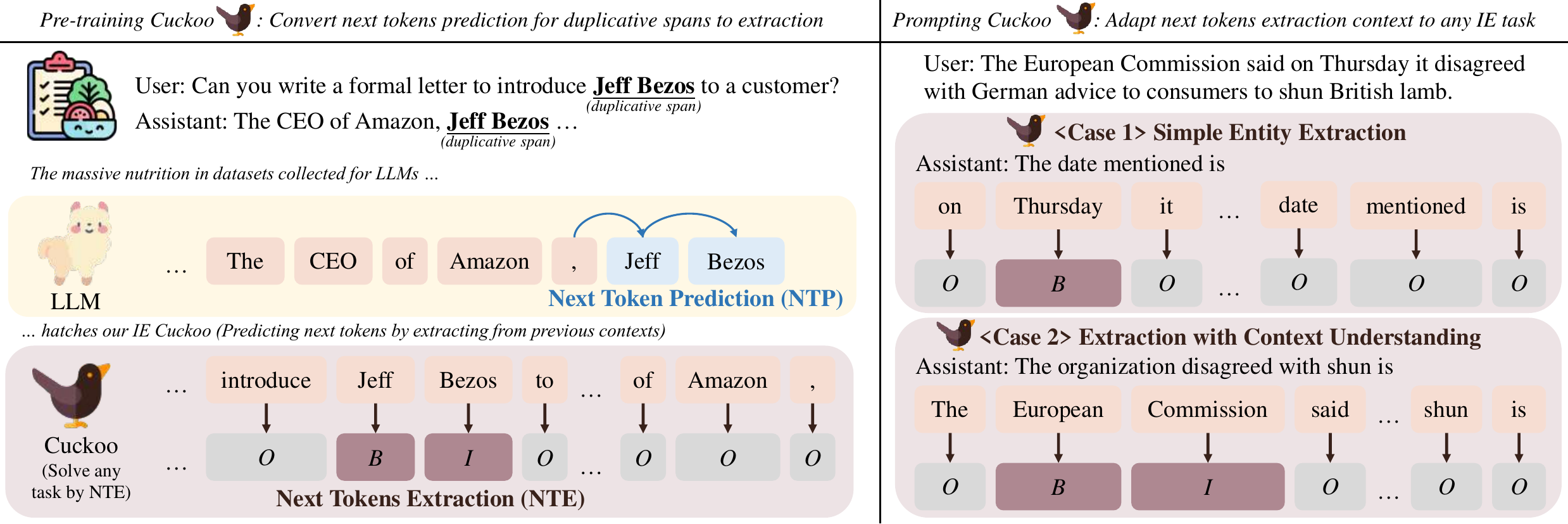}
    \caption{\our takes a free ride on LLM resources (e.g., C4 and TuluV3~\citep{tulu3}) by formalizing next token prediction for duplicative spans as extraction in the BIO paradigm. During the inference, the prompts can be adjusted to different extractive tasks, making \our a versatile IE model.
    }
    \label{fig:paradigm}
    \vspace{-5mm}
\end{figure*}

This paper proposes a frustratingly simple yet effective way to scale up IE pre-training. We suggest that IE pre-training can simply be a free rider on the LLM's training resources by learning on exactly the same pre-training and post-training datasets. We modify NTP to next tokens extraction (NTE), using BIO tags for next tokens that can be extracted from the input context as shown in Figure~\ref{fig:paradigm}. With the instruction-following ability learned in post-training, one can adjust the prompt to instruct NTE-based taggers to perform different IE tasks.

Specialized for IE, NTE has three advantages over NTP. 1) Parameter Efficiency, NTP requires extra parameters to store knowledge to generate tokens not in the input context, while NTE concentrates only on tagging input tokens. 
Thus, NTE-based IE taggers can have better parameter efficiency than NTP-based LLMs, fitting it to smaller models like RoBERTa~\citep{roberta}. 2) Inference Efficiency, NTE taggers are not only smaller because of the parameter efficiency but can also extract multiple tokens with the BIO scheme in one forward pass. 3) Transferability, NTE taggers can easily adapt to IE tasks, which are typically annotated in the same BIO scheme.

\begin{figure}
    \centering
    \includegraphics[width=\linewidth]{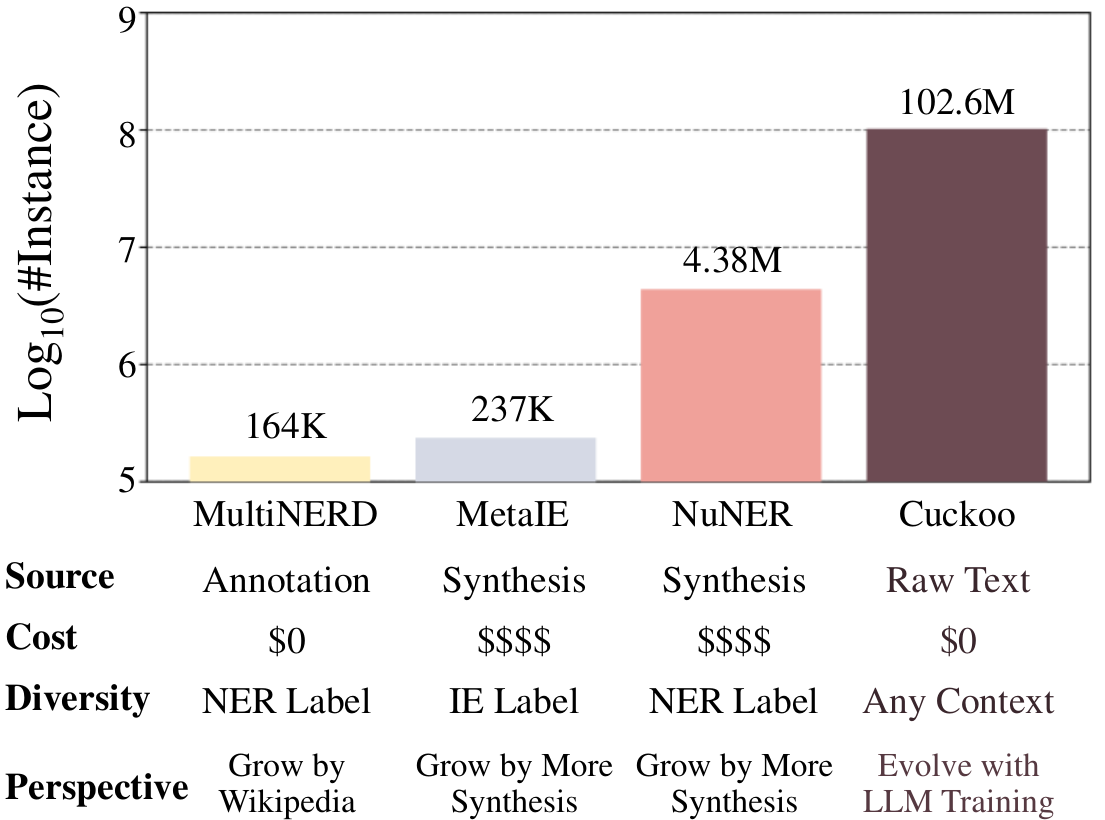}
    \caption{Comparison of scale, cost, and diversity among different IE pre-training datasets. Our data collection for \our is free by converting LLM's learning resources, which forces the tagger to learn from diverse contexts. \our can also evolve with the data collection for LLM's post-training.}
    \label{fig:cuckoo_data_scale}
    \vspace{-5mm}
\end{figure}

With NTE, we easily collect $100$M pre-training instances from C4\footnote{We estimate the English part of C4 can be transformed into $5$B instances, we only take $100$M $(2\%)$ for experiment efficiency.}~\citep{t5}, a popular pre-training dataset, and $2.6$M chat-formatted instances from TuluV3 post-training dataset~\citep{tulu3} to endow the model with instruction-following ability.
We continually train a RoBERTa tagger on massive NTE data, which results in our \emph{\our} model, a free rider with a training paradigm similar to NTP on training resources for LLMs. We present the comparison of scale, cost and diversity with other IE pre-training datasets in Figure~\ref{fig:cuckoo_data_scale}.

We follow the few-shot adaptation evaluation in previous works~\citep{multinerd,NuNER} to benchmark \our, which shows that \our is as versatile as LLMs in extractive tasks. Training with few-shot data, \our can quickly understand different kinds of NER labels, free text questions in machine reading comprehension, and complex instructions, to perform precise extraction. With overwhelming advantages in data scale, \our outperforms models pre-trained on massive human-annotated or LLM-synthesized datasets by a large margin.

Finally, we analyze to show 1) \our can evolve with the data collection for LLM's post-training data; 2) in-context tagging ability emerges in \our just like in-context learning in LLMs; and 3) \our scales up by the increasing number of our constructed NTE data.

%% file: 2-rel.tex
\section{Background}

\paragraph{Information Extraction} Information extraction (IE) is one of the most fundamental applications in natural language processing. IE systems take the user's requirement (e.g., defined by a label text, a question, or an instruction) and extract spans of several tokens from input texts. The two most frequent categories of IE targets are entity and relation, which structure many IE tasks, such as named entity recognition~\cite{conll2003}, relation extraction~\cite{conll2004}, event extraction~\citep{ace2005multilingual}, and others~\citep{srl-task,DBLP:conf/semeval/PontikiGPPAM14,aste-task}. A crucial challenge to modern IE systems is the growing number of IE targets (e.g., various label names) in the open world, which are scarce in annotation and require IE systems for quick transfer learning. Thus, many works have collected massive automated IE annotations to pre-train IE models~\cite{fewnerd,multinerd,TadNER,NuNER,metaie}, which shows benefits in transferring to low-resource IE targets.

\paragraph{Large Language Model} The biggest game-changer for natural language processing in all domains is the large language model (LLM)~\citep{tulu,llama-2,achiam2023gpt4,olmo,dubey2024llama3,team2024gemma}. Learning on trillions of tokens for pre-training and post-training, LLMs have shown surprisingly strong performance on all kinds of tasks~\citep{achiam2023gpt4}. Next token prediction, the paradigm behind the success of LLMs, supports exploiting every token in raw texts as the annotation to strengthen the model's capability. Consequently, many IE researchers have turned toward LLMs~\citep{llm4clinicalie,gpt-ner,llm4ie} to use them as strategic information extractors with planning~\citep{LLM_Plan,LLM_NestNER} and chain-of-thoughts~\citep{chain_of_thoughts,cot_re}.

\paragraph{Pre-training Paradigm: IE v.s. LLM} The rise of LLMs has challenged the meaningfulness of IE pre-training with an overwhelmingly larger number of annotations. The lagging of IE pre-training can be attributed to the relatively high format requirement for IE annotation like labels in Wikipedia links. This paper shows IE pre-training can take a free ride on LLM's NTP paradigm to unleash the power of massive pre-training.

%% file: 3-method.tex
\section{Our \our}

\subsection{Next Tokens Extraction}

The learning paradigm for LLMs is next token prediction (NTP), which calculates the representation of a context $[x_1, x_2, \cdots, x_t]$ to output a probability distribution $p_{t+1}$ of the next token $x_{t+1}$ over all potential tokens in the LLM's vocabulary. The prediction $p_{t+1}$ is optimized by the cross entropy loss to maximize its value on $x_{t+1}$.

We modify NTP into next tokens extraction (NTE) for cases that the span of next $n$ tokens $[x_{t+1}, \cdots, x_{t+n}]$ already exist in the context $[x_1, x_2, \cdots, x_t]$, such that $[x_{k+1}, \cdots, x_{k+n}] = [x_{t+1}, \cdots, x_{t+n}] (1 \leq k \leq t-n)$. When we detect such $(t,k,n)$, we annotate IE tags for the context as $[l_1, l_2, \cdots, l_t]$ following a BIO scheme. We first set all tags $l$ to \textit{O}. As there can be multiple $k$ for $t$, for each $k$, we set $l_k$ to \textit{B} and $[l_{k+1}, \cdots, l_{k+n}]$ to \textit{I}. The high-level idea of NTE is to replace prediction by extraction for duplicative spans that appear multiple times in the context.

NTE thus allows IE pre-training to directly exploit NTP datasets for LLM training, which significantly broadens the potential training data. During the inference, one can adjust the prompts of an NTE-based tagger to instruct it to perform different kinds of extractive tasks. Recall the strengths mentioned for NTE in the introduction, NTE specialized for IE has advantages in parameter efficiency, inference efficiency, and adaptability over NTP. 

\subsection{Massive Nutrition for \our}

\paragraph{Pre-training and Post-Training} With NTP-to-NTE conversion, we can simply copy the two training stages for LLMs, to perform pre-training and post-training for NTE-based IE taggers. Pre-training learns raw texts while post-training learns instruction-following dialogues between the user and the IE assistant. During pre-training, we annotate BIO tag sequences based on all $(t,k,n)$ triplets, assuming the multiple appearances of the same span of tokens indicate a certain level of extractive relation~\citep{UCPhrase}. For post-training, we suppose the extraction should focus on the texts provided by users so we only keep $(t,k,n)$ triplets that $k$ falls in the user's request and $t$ falls in the assistant's response.

Then, we select the resources for pre-training and post-training. While the NTE framework allows us to exhaust all kinds of resources, we use only one dataset for each stage for experiment efficiency. For pre-training, we select the popular C4 (CommonCrawl) dataset~\citep{t5}, which contains $4$B passages and is commonly used to pre-train LLMs. For post-training, we use the most advanced TuluV3~\cite{tulu3} dataset with $939$K instruction-following interactions between the user and the assistant.

To further boost the experiment efficiency, we first collect noun phrases parsed by SpaCy\footnote{\href{https://spacy.io/}{https://spacy.io/}}, filtering stop words or punctuations. Then we collect $5\%$ of the rest spans (no overlapping) that are duplicative to produce NTE instances. On C4, we keep the first $100$M NTE instances transformed from the raw texts. On TuluV3, we transform all post-training interactions into the NTE format, resulting in $2.6$M instances. We also sample $5\%$ spans not existing in their previous contexts, whose NTE labels are annotated by all \textit{O} as negative cases.

With the $102.6$M instances, we continually pre-train a \texttt{roberta-large} model~\citep{roberta} as the BIO tagger for NTE, optimized by AdamW~\citep{AdamW} with learning rate initialized to $10^{-5}$. The batch size is set to $64$, taking about $1.6$M steps for the optimization.


\subsection{Statistics}

Besides the huge scale, we analyze other key statistics of our massive NTE dataset to investigate its efficiency in learning various IE targets. Our investigation is respectively done on the two pre-training and post-training data splits. 

\paragraph{How ``extractive'' are the data?} An obvious concern on the NTE dataset is whether the automated annotations reflect real extractive relations. We prompt the advanced LLM, \texttt{gpt-4o}~\citep{achiam2023gpt4}, to identify whether NTE data establish real extractive relations. The responses on $20$K sampled data show $93.39\%$ pre-training data and $96.20\%$ post-training data contain extractive relations, which shows the high data efficiency of the annotation strategy.

\paragraph{How diverse are the data?} The data is extremely diverse by containing any duplicative spans in a broad domain. We find around $28$M unique spans in C4 and $0.4$M in TuluV3, which is combined with highly diverse contexts in C4 and TuluV3. Our dataset covers various span lengths (maximally $40$ words) and context lengths (maximally $512$ words). The proportion of span with $\geq 4$ tokens is $4.52\%$, which seems small but still contains $4.6$M spans because of the large scale of our dataset. Our context length is also more diverse than previous IE pre-training resources~\citep{multinerd,NuNER,metaie} where data only have one or two sentences as context.

\paragraph{What is the conversion rate?} The conversion rate from a sentence to an NTE instance is $332\%$ for C4 and $235\%$ for TuluV3. This is highly efficient in comparison with traditional IE pre-training datasets relying on scarce links or expensive synthesis. The full C4 dataset can be transformed into $5$B NTE instances. However, the efficiency is still relatively lower than NTP. Only $4.06\%$ tokens in pre-training and $4.14\%$ tokens in post-training are used for NTE tagger learning, which indicates the supervision from LLM resources can be further augmented.

%% file: 4-exp.tex
\section{Experiments}

\begin{table}
\centering
\small
\scalebox{1.0}{
\begin{tabular}{p{1.2cm}p{5.5cm}}
\toprule
Level & Example\\
\midrule
Basic &  Organization\\
\midrule
Query & Which organization launched the campaign?\\
\midrule
\multirow{3}*{Instruction} & Organization (Disambiguation: The organization entity must be a subject of any active action in the context.)\\
\bottomrule
\end{tabular}
}
\caption{IE targets of different understanding levels.} 
\vspace{-5mm}
\label{tab:level}
\end{table}

Different from previous evaluation procedures that enumerate IE tasks~\citep{UIE,tanl,metaie}, our evaluation splits IE tasks into different levels of understanding the IE target. Specifically, the three levels are 1) Basic IE, understanding a single label text for an entity or a relation, such as named entity recognition. 2) Query-based IE, understanding a sentence-level query, such as machine reading comprehension (MRC). 3) Instruction-following IE, understanding complex extractive instructions like LLMs.

Examples of different understanding level are enumerated in Table~\ref{tab:level}. We expect that \our will be comparable to traditional IE pre-training on Basic IE as most popular label texts have been enumerated by LLM synthesis~\citep{NuNER,metaie}. \our's advantage over traditional IE pre-training is on query-based and instruction-following IE, which requires understanding more complex IE targets.

\subsection{Benchmark and Evaluation}

Following the high-level evaluation objective, we use several traditional benchmarks for each level of IE ability. Method and benchmark details are included in Appendices~\ref{apdx:detail} and~\ref{apdx:itie}.

\paragraph{Basic IE} benchmarks the understanding of simple labels for entity and relation. We include $4$ named entity recognition datasets (CoNLL03~\citep{conll2003}, BioNLP2004~\citep{bionlp2004}, MIT-Restaurant/Moive~\citep{tner}) and $2$ relation extraction datasets (CoNLL04~\citep{conll2004} and ADE~\citep{ADE}).

\paragraph{Query-based IE} requires the understanding of more complex sentence-level semantics of the IE target. We thus include $3$ machine reading comprehension datasets (SQuAD~\citep{SQuAD}, SQuAD-V2~\citep{SQuAD-V2}, DROP~\citep{DROP}). We filter out non-extractive questions in DROP.

\paragraph{Instruction-following IE} is a feature of LLMs when they are applied for IE. Users can include detailed requirements for the IE target in the prompt, which is hard for traditional IE systems that only understand simple label texts. However, instruction-following IE currently lacks of benchmarks\footnote{Existing InstructIE benchmarks~\citep{InstructIE_UIUC,InstructIE_ZJU} concentrate more on using instruction for traditional IE than instruction-awareness.}. Based on the real role of instruction in IE, we apply rules and a strong LLM, \texttt{GPT-4o}, to synthesize $3$ instruction-following IE by modifying traditional benchmarks. 1) \textbf{Disambiguation}, we write a definition instruction for $3$ ambiguous types, (\textit{``Organization''} in CoNLL2003, \textit{``Protein''} in BioNLP2004, \textit{``Location''} in MIT-Restaurant), such as \textit{``Disambiguation: The organization entity must be a subject of any active action in the context.''}. We use \texttt{GPT-4o} to filter out entities that no longer meet the IE target, resulting in a new instruction-following IE benchmark. 2) \textbf{Preference}, there are different ground truth answers in machine reading comprehension like ``Bruno Mars'', ``Mars''. However, one might prefer the longer or the shorter answer. Thus, we modify the SQuAD dataset with $3$ instructions with a preference for ``Longer answer'', ``Shorter answer'', ``Concise answer (Answer with no extra words)''\footnote{This means when ``Los Angeles'', ``the US'' and ``US'' all exist in the answer candidates, ``the US'' will be removed but ``Los Angeles'' will be kept.}. This filtering modification is automated by functions with no LLM involved. 3) \textbf{Miscellaneous}, we write $3$ instructions to define the \textit{``Miscellaneous''} entity type in CoNLL2003, MIT-Restaurant, and MIT-Movie. In practice, we clarify the existing miscellaneous type for CoNLL2003 and combine $3$ minority types as miscellaneous for MIT-Restaurant and MIT-Movie. We calculate metrics only on miscellaneous entities to evaluate whether the model can understand the scope definitions.

The evaluation continues with the model's few-shot adaptability. The model will be fine-tuned on a few examples in the training set and then evaluated on the test set. For basic IE, we will have $5$ shots for each entity/relation category. For query-based IE, we will have $32$ training examples. For instruction-following IE, the definition of few-shot follows the original dataset. We include more details for the construction of instruction-following IE benchmark in Appendix~\ref{apdx:itie}.

We benchmark IE performance with the traditional F1 score. For Basic IE, it refers to the Micro F1 for labeled entity spans. In Query-based IE, the F1 score refers to the maximal word-level F1 between the answer and one of the ground truths. Instruction-following IE benchmarks follow the metric of the original datasets.


\subsection{Baselines and Variants}

\input{table_basic_ie}

We incorporate baselines into our experiments to validate our two main claims. 1) NTE is a paradigm that can scale up the data resources for IE pre-training, which learns taggers with better few-shot adaptability, especially in instruction-following. 2) NTE is a more efficient paradigm than NTP for IE, which results in significantly stronger extractive ability of NTE-based taggers than NTP-based LMs.

For 1), we include previous IE pre-training resources to compare their pre-training effects with our NTE-based dataset. These resources include,

\begin{itemize}[nosep,leftmargin=*]
    \item \textbf{MultiNERD}~\citep{multinerd} is a NER pre-training dataset based on Wikipedia and Wikinews, which contains $164$K instances in the English split with $17$ label names. The annotations are from community contributors.
    \item \textbf{NuNER}~\citep{NuNER} is a massive NER pre-training dataset synthesized by ChatGPT-$3.5$~\citep{chatgpt} on massive raw texts. NuNER has $4.38$M instances with $273$K unique label names.
    \item \textbf{MetaIE}~\citep{metaie} is a massive IE pre-training dataset synthesized by ChatGPT-$3.5$ and $4$ with a broader coverage than simple NER. The LLMs are prompted to enumerate possible important information for entities and relations. MetaIE includes $237$K IE instances with $31$K unique label names.
\end{itemize}

In addition to resources using annotations for label names, we also consider machine reading comprehension as a pre-training task for IE, as it can be viewed as query-based IE. We thus include,

\begin{itemize}[nosep,leftmargin=*]
    \item \textbf{MRQA}~\citep{MRQA} is a collection of machine reading comprehension data that extracts an answer from a passage for a question in each instance. We exclude SQuAD as it is used for benchmarking, which remains $488$K instances.
\end{itemize}

For 2), we use the same resources for \our (C4+TuluV3) to continually pre-train an OPT model~\citep{OPT} in the same parameter scale ($\sim300$M) as the base model RoBERTa of \our. We select OPT because its NTP pre-training resource has covered the one for RoBERTa~\citep{roberta,OPT}, which eliminates the attribution of \our's advantage to a better base model (RoBERTa).

For the ablation study, we include the variants of \our, which only use the LLM's pre-training (C4) or post-training (TuluV3) resource for IE pre-training. These two variants aim to demonstrate the contributions of both stages to justify the imitation of the LLM's training pipeline. 


\paragraph{Rainbow \our} Finally, we incorporate a strong variant combining more post-training resources, \emph{Rainbow \our}. Rainbow \our extends the post-training resource from only TuluV3 to merging multiple datasets including samples from MultiNERD, NuNER, MetaIE, and MRQA, which aims to exploit all possible resources to further boost the IE pre-training.

\paragraph{Zero-shot Performance} is also evaluated on our \our and its variant Rainbow \our to demonstrate the direct performance after the IE pre-training on LLM's resources.

\paragraph{Comparison with LLMs} is discussed in Appendix~\ref{apdx:vs_llm} to expand the comparison scope.

\subsection{Basic IE}

The performance on basic IE tasks is presented in Table~\ref{tab:basic_ie}. Our two main claims are supported by the experiment results,

1) \our outperforms all baselines using different IE pre-training resources on both entity and relation extraction. Among the baselines, the best-performing ones are NuNER for entity and MRQA for relation, which they specialize in. \our overwhelms the baselines with a much larger pre-training data scale. As \our with only the raw texts from C4 (pre-training) has already achieved comparable or better performance than baselines, the conversion to NTE shows strong data efficiency on raw texts. 

2) The NTE pre-trained RoBERTa (\our) outperforms the NTP pre-trained OPT, which validates our intuition that language models can be more parameter efficient by focusing on extraction.

Besides the validation of our main claims, we also have more discoveries from the performance of variants. The first observation is that both pre-training and post-training datasets contribute to adaptability. In basic IE tasks, the massive raw texts in C4 contribute more than the curated post-training data in TuluV3, which indicates the basic IE tasks are simple enough to be well transferred by learning without annotations. The Rainbow \our shows \our can be further enhanced with merging more post-training resources, demonstrating significantly strong IE ability. 

\subsection{Query-based IE}

\input{table_query_ie}

We present the performance of models on query-based IE (MRC) in Table~\ref{tab:query_ie}. Among out-of-domain models, \our significantly outperforms other models pre-trained on basic IE tasks, rivaling the model pre-trained on the in-domain MRQA dataset. The result exhibits the benefit of NTE to pre-train in a wild and diverse raw text distribution, contrasting the fixed templates in basic IE pre-training. Post-training resources show a more significant contribution to query-based than basic IE tasks as queries in MRC require higher instruction awareness. Merging MRQA into the pre-training, Rainbow \our shows a significant advantage over using only MRQA via unifying all kinds of pre-training resources by the NTE paradigm. 

\subsection{Instruction-following IE}

\input{table_instruct_ie}

Table~\ref{tab:instruct_ie} demonstrates the instruction-following ability of different IE models. The zero-shot performance implies that the task requires a higher-level understanding of IE instructions. \our once again significantly outperforms other models except for an in-domain case (MRQA on MRC-based preference instruction testing) and widens the gap, showing its strong adaption to new instructions with the following ability learned from LLM pre-training resources. Post-training data contribute the most to the ability to follow instructions, playing the same role as for LLMs. Occasionally, learning only post-training data outperforms the full \our. Rainbow \our, with a large amount of post-training supervision, once again significantly boosts the performance.

\begin{table}
\centering
\small
\resizebox{\linewidth}{!}
{
\begin{tabular}{lcccc}
\toprule
Method & {Long} & {Short} & {AnsSim $\downarrow$} & {DualEM} \\
\midrule
\our & $57.84$ & $51.39$ & $\textbf{40.48}$ & $11.67$ \\
MRQA & $62.61$ & $61.05$ & $48.17$ & $12.32$ \\
Rainbow \our & $\textbf{67.20}$ & $\textbf{63.67}$ & $44.58$ & $\textbf{18.95}$ \\
\bottomrule
\end{tabular}
}
\caption{Detailed analysis on the instruction-following ability of IE models with preference as an example.}
\vspace{-3mm}
\label{tab:instruct_ability}
\end{table}

\paragraph{\our reacts to instruction.} We provide a deeper investigation of \our's reactions to instructions. Specifically, we test the preference instructions for the longest and shortest answers, which will lead to different answers. We fine-tune pre-trained IE models with few shots for both the longest and the shortest answers and then test their instruction-following ability. For evaluation, we use answer similarity (AnsSim) between outputs from two instructions, where higher similarity indicates less instruction-awareness. We also use dual exact matching (DualEM) as a strict metric to evaluate whether the model correctly reacts to both instructions. AnsSim calculates the word-level F1 score between answers from two instructions and DualEM refers to the model accuracy to produce both answers correctly. Table~\ref{tab:instruct_ability} shows that the MRQA model is no longer significantly better than \our on DualEM. AnsSim also indicates MRQA model to have less instruction-awareness, restraining its strong MRC ability to be applied with specific instructions. In comparison, the Rainbow \our shows a much higher advantage over the MRQA model according to the DualEM metric, demonstrating a better efficiency in applying the MRC ability to the instruction-following scenario. 


%% file: table_basic_ie.tex
\begin{table*}
\centering
\small
\resizebox{\linewidth}{!}{
\begin{tabular}{llcccccccc}
\toprule
\multicolumn{2}{l}{\multirow{2}*{Method}} & \multicolumn{5}{c}{Named Entity Recognition} & \multicolumn{3}{c}{Relation Extraction} \\
\cmidrule(l){3-7} \cmidrule(l){8-10} 
& & CoNLL2003 & BioNLP2004 & MIT-Restaurant & MIT-Movie & Avg. & CoNLL2004 & ADE & Avg. \\
\midrule
\multirow{2}*{\rotatebox{90}{zero}}& \our & $35.38$ & $23.62$ & $8.11$ & $9.06$ & $19.04$ & $48.95$ & $34.67$ & $41.81$ \\
& Rainbow \our & $38.56$ & $22.07$ & $35.38$ & $29.53$ & $31.39$ & $53.81$ & $62.01$ & $57.91$ \\
\midrule
\multirow{10}*{\rotatebox{90}{few-shot}}& OPT-C4-TuluV3 & $50.24$ & $39.76$ & $58.91$ & $56.33$ & $50.56$ & $47.14$ & $45.66$ & $46.40$ \\
& RoBERTa & $33.75$ & $32.91$ & $62.15$ & $58.32$ & $46.80$ & $34.16$ & $2.15$ & $18.15$ \\
& MRQA & $72.45$ & $55.93$ & $68.68$ & $66.26$ & $65.83$ & $66.23$ & $67.44$ & $66.84$ \\
& \our & $73.60$ & $57.00$ & $67.63$ & $67.12$ & $\textbf{66.34}$ & $69.57$ & $71.70$ & $\textbf{70.63}$ \\
& $\quad$ Only Pre-train & $72.46$ & $55.87$ & $66.87$ & $67.23$ & $65.61$ & $68.14$ & $69.39$ & $68.77$ \\
& $\quad$ Only Post-train & $72.80$ & $56.10$ & $66.02$ & $67.10$ & $65.51$ & $68.66$ & $69.75$ & $69.21$ \\
\cmidrule(lr){2-10}
& MultiNERD$^\dag$ & $66.78$ & $54.62$ & $64.16$ & $66.30$ & $60.59$ & $57.52$ & $45.10$ & $51.31$ \\
& NuNER$^\dag$ & $74.15$ & $56.36$ & $68.57$ & $64.88$ & $65.99$ & $65.12$ & $63.71$ & $64.42$ \\
& MetaIE$^\dag$ & $71.33$ & $55.63$ & $70.08$ & $65.23$ & $65.57$ & $64.81$ & $64.40$ & $64.61$ \\
& Rainbow \our$^\dag$ & $79.94$ & $58.39$ & $70.30$ & $67.00$ & $\textbf{68.91}$ & $70.47$ & $76.05$ & $\textbf{73.26}$ \\
\bottomrule
\end{tabular}
}
\caption{Performance comparison on Basic IE Tasks.  $\dag$: In-domain Transfer. (Transfer learning on the same task and format as the pre-training stage.)}
\label{tab:basic_ie}
\end{table*}

%% file: table_query_ie.tex
\begin{table}
\centering
\small
\resizebox{\linewidth}{!}{
\begin{tabular}{llcccc}
\toprule
\multicolumn{2}{l}{Method} & SQuAD & SQuAD-V2 & DROP & Avg. \\
\midrule
\multirow{2}*{\rotatebox{90}{zero}} & \our & $48.82$ & $49.16$ & $38.41$ & $45.46$ \\
& Rainbow \our & $82.79$ & $57.67$ & $61.62$ & $67.36$ \\
\midrule
\multirow{10}*{\rotatebox{90}{few-shot}}& OPT-C4-TuluV3 & $39.80$ & $53.81$ & $31.00$ & $41.54$ \\
& RoBERTa & $31.86$ & $48.55$ & $9.16$ & $29.86$ \\
& MultiNERD & $42.85$ & $50.99$ & $30.12$ & $41.32$ \\
& NuNER & $61.60$ & $52.67$ & $37.37$ & $50.55$ \\
& MetaIE & $74.59$ & $62.54$ & $30.73$ & $55.95$ \\
& \our & $77.47$ & $64.06$ & $54.25$ & $\textbf{65.26}$ \\
& $\quad$ Only Pre-train & $75.64$ & $63.36$ & $52.81$ & $63.94$ \\
& $\quad$ Only Post-train & $77.05$ & $62.39$ & $54.80$ & $64.75$ \\
\cmidrule(lr){2-6}
& MRQA$^\dag$ & $80.07$ & $66.22$ & $54.46$ & $66.92$ \\
& Rainbow \our$^\dag$ & $86.57$ & $69.41$ & $64.64$ & $\textbf{73.54}$ \\
\bottomrule
\end{tabular}
}
\caption{Performance comparison on Query-based IE Tasks.
$\dag$: In-domain Transfer.
}
\vspace{-3mm}
\label{tab:query_ie}
\end{table}

%% file: table_instruct_ie.tex
\begin{table}
\centering
\small
\resizebox{\linewidth}{!}{
\begin{tabular}{llcccc}
\toprule
\multicolumn{2}{l}{{Method}} & {Disamb.} & {Prefer.} & {Misc.} \\
\multicolumn{2}{l}{{Base Task}} & {NER} & {MRC} & {NER} \\
\midrule
\multirow{2}*{\rotatebox{90}{zero}} & \our & $13.88$ & $35.56$ & $2.93$ \\
& Rainbow \our & $21.93$ & $60.81$ & $14.62$ \\
\midrule
\multirow{10}*{\rotatebox{90}{few-shot}}& OPT-C4-TuluV3 & $28.56$ & $53.68$ & $37.19$ \\
& RoBERTa & $12.29$ & $6.04$ & $9.71$ \\
& MultiNERD & $31.71^\dag$ & $30.84$ & $44.68^\dag$  \\
& NuNER & $31.40^\dag$ & $51.01$ & $44.32^\dag$ \\
& MetaIE & $29.77^\dag$ & $56.12$ & $47.35^\dag$ \\
& \our & $\textbf{34.97}$ & $62.53$ & $\textbf{49.17}$ \\
& $\quad$ Only Pre-train & $32.21$ & $59.64$ & $46.05$ \\
& $\quad$ Only Post-train & $34.28$ & $\textbf{64.37}$ & $47.28$ \\
& MRQA & $29.33$ & $66.83^\dag$ & $48.67$ \\
& Rainbow \our & $\textbf{37.75}^\dag$ & $\textbf{70.95}^\dag$ & $\textbf{51.86}^\dag$ \\
\bottomrule
\end{tabular}
}
\caption{Performance comparison on Instruction-following IE tasks for disambiguation (Disamb.), preference (Prefer.), and miscellaneous (Misc.).
$\dag$: In-domain Transfer.}
\label{tab:instruct_ie}
\end{table}

%% file: 5-analysis.tex
\section{Analyses}

\subsection{Evolution with LLMs}

\begin{figure}
    \centering
    \includegraphics[width=\linewidth]{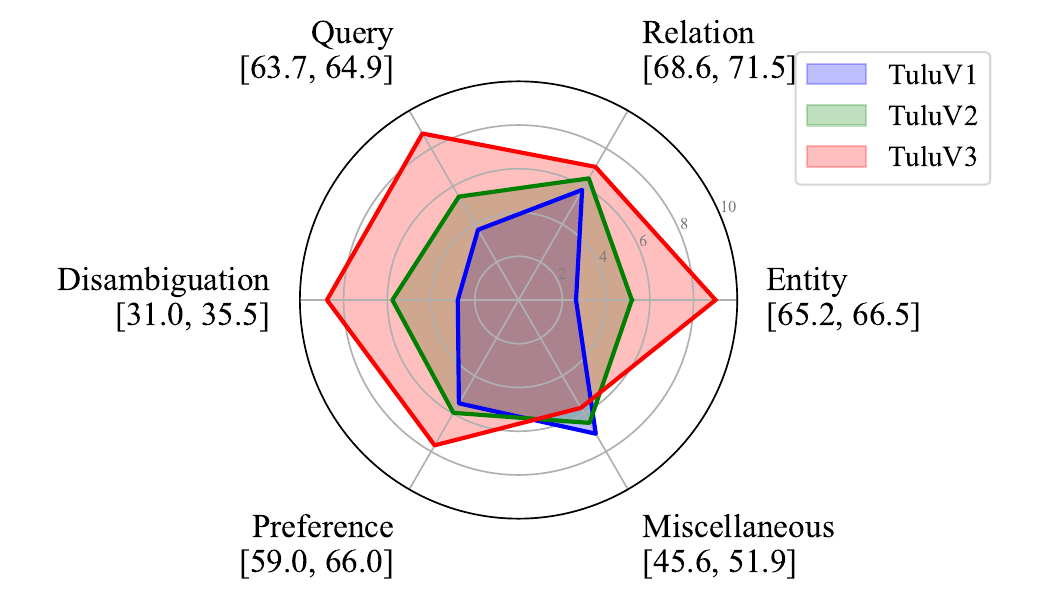}
    \caption{The evolution of Cuckoo with LLM's post-training resources. Domain $[\mu-2\sigma, \mu+2\sigma]$ is annotated under each evaluation dimension.}
    \label{fig:coevolution_radar}
\end{figure}

A feature of our \our is its evolution with LLM's training resources, especially for post-training data which are progressively curated by researchers~\citep{olmo,wizard_lm,tulu3}. In Figure~\ref{fig:coevolution_radar}, we plot the performance of \our post-trained by different versions of Tulu post-training datasets from V1 to V3~\citep{tulu,tulu2,tulu3} after pre-training on C4. All performances are normalized by a linear mapping from $[\mu-2\sigma, \mu+2\sigma]$\footnote{$\mu, \sigma$ are based on the performance of $4$ \our models (before post-training, after post-training with TuluV1 to V3)} to $[0, 10]$ for demonstration. The result illustrates a evolution between \our and the LLMs. With each evolution in post-training data collection for LLMs, \our's performance can also be expanded in most dimensions. In the future, \our can be further improved together with the quality of LLM's training data with the free-riding feature of our NTE paradigm. 

\subsection{Emergence of In-context Tagging}

\begin{figure}
    \centering
    \includegraphics[width=\linewidth]{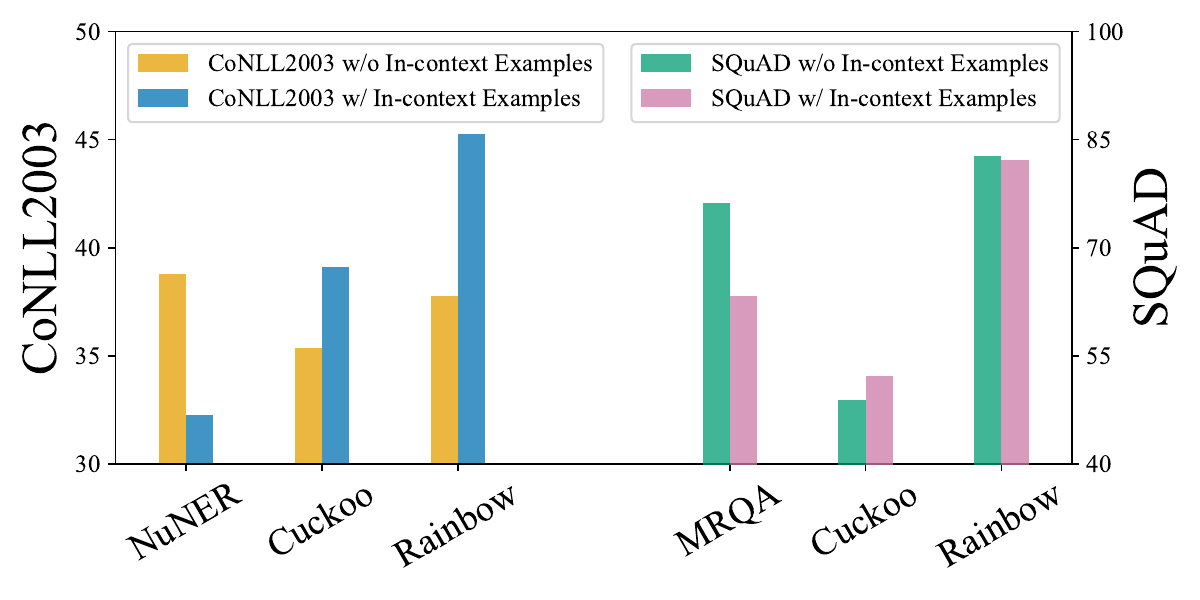}
    \caption{In-context tagging ability emerges in Cuckoo but not in IE models pre-trained by other resources.}
    \label{fig:ict}
\end{figure}

In-context learning is an emerging skill in LLMs that adapts LLMs to new tasks with examples in the given context. We investigate whether in-context learning appears in \our, which uses a similar learning paradigm and resource as LLMs. We append $5$ examples for CoNLL2003 and $1$ example for SQuAD (due to context window limitation) to the context and test the in-context tagging performance of different models. In Figure~\ref{fig:ict}, we find only \our able to improve (at least retain) its IE ability while other models (even pre-trained on similar tasks) show a significant drop. Thus, NTE on LLM's resources is verified to enable in-context tagging for \our. As suggested in~\citet{emergence_of_icl}, the occasional burstiness in raw texts contributes to the emergence of in-context tagging in \our. While NuNER and MRQA are well formalized, they fail to learn models with in-context learning ability because of the lack of burstiness.

\subsection{Data Scaling Trend}

Data is an important factor in the scaling law~\citep{scaling_law}. Thus, we test the transfer learning ability of checkpoints pre-trained with different data scales to downstream tasks. We focus on the scaling law of raw texts in C4 as they are cheaper to scale up and we have discussed the evolution of \our with post-training data collection. Our investigation covers both early pre-training stages to $4.1$M instances and the scaling-up to $100$M.

\begin{figure}
    \centering
    \includegraphics[width=0.99\linewidth]{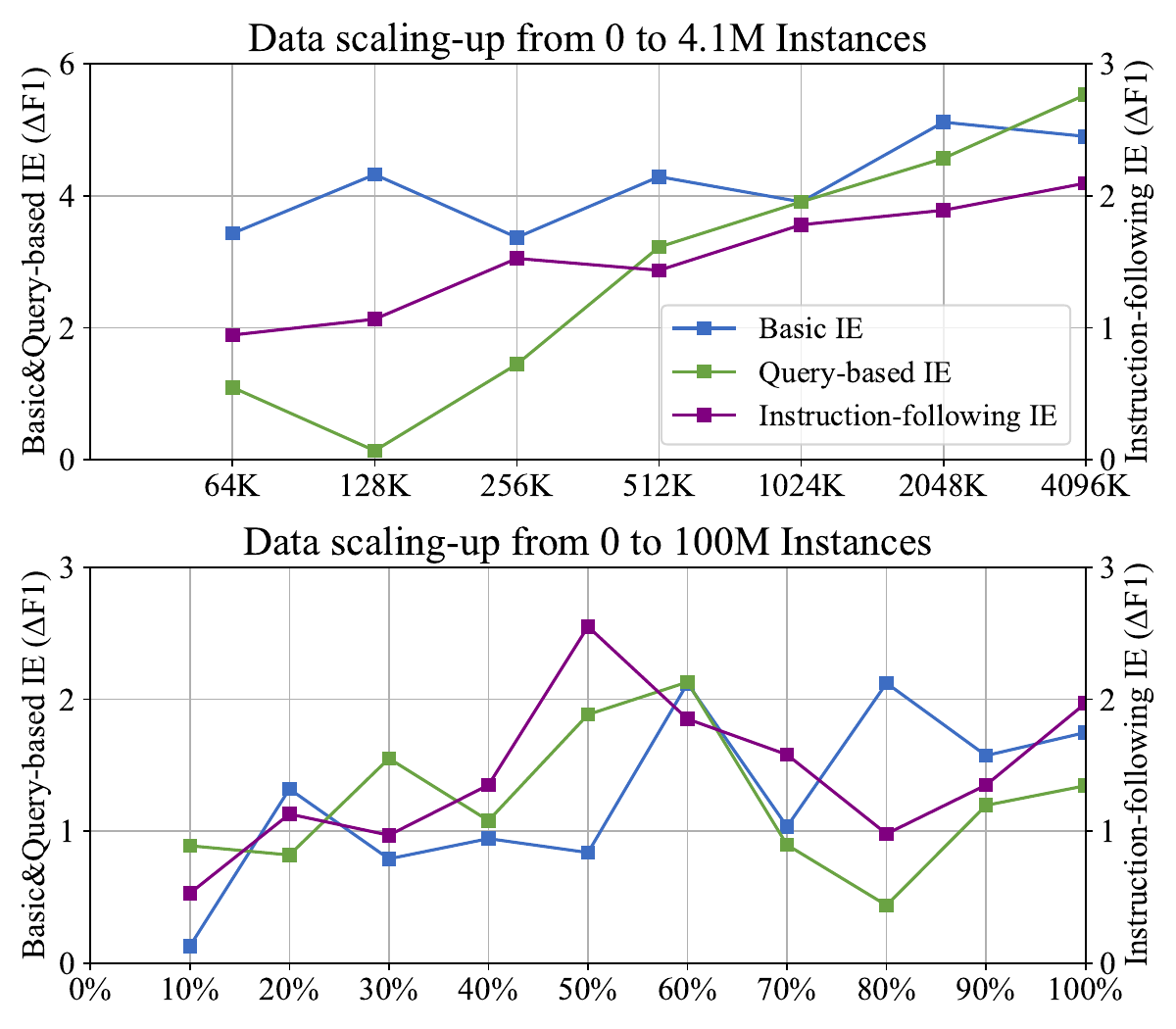}
    \caption{The data scaling trend of \our on the early $4.1$M C4 instances and the massive $100$M instances.}
    \label{fig:scale}
\end{figure}

In the two subfigures of Figure~\ref{fig:scale}, we plot the data scaling trend in pre-training \our.  The upper figure shows a clear performance rising trend together with the increasing data amount, indicating all dimensions of IE ability are scaled-up in the early pre-training stage. In the scaling-up to $100$M stage, the macroscopic trend retains its steady increase but turbulence emerges. Some intermediate checkpoints like at $50\%\sim60\%$ data scale show a competitive performance with the fully pre-trained model. This implicates that the capacity of the small RoBERTa might meet its bound, and further improvement requires more parameters.

%% file: 6-con.tex
\section{Conclusion and Future Work}

This paper proposes a large-scale IE pre-training paradigm with the LLM's pre-training and post-training resources. The massive nutrition incubates a versatile \our model, which outperforms the pre-training with previous IE resources. \our can evolve with the data preparation for LLMs.
Further work on \our will focus on variants in learning paradigms, datasets, and backbones.

%% file: 7-lim.tex
\section*{Limitations}

While \our validates the strength of NTE to take a free ride with LLM resources, our scope can be extended to several topics out of the main claims.

\paragraph{Label Embedding} Some IE paradigms (e.g., original NuNER) learns label embeddings to efficiently label the extracted spans. As \our imitates NTP to perform NTE, its IE process requires enumerating the label names similar as the generative IE using LLMs. Matching label embedding has its efficiency advantage while generative IE allows the label texts to interact with the context, resulting in potentially better performance. \our follows the generative IE paradigm to pursue better performance based on the established success of LLMs. However, future efforted can be devoted into a label embedding version of \our, which takes the context as the label text to boost the IE efficiency.

\paragraph{Data Source} The C4 corpus for raw text features broad coverage. However, recent progress in LLMs shows that specific sources of pre-training data (e.g., textbooks) benefit certain skills of LLMs, such as math. This paper only discusses C4 to avoid the IE performance improvement attributed to a specific data source. Future works can extend our scope to compare the effect of all kinds of resources in pre-training, which might find certain resources are superior in IE pre-training using NTE.

\paragraph{Backbone Variants} The current scopes is designed to justify the benefit of NTE in gathering massive IE pre-training data. Thus, the comparison is biased to data quality rather than backbone models. Further exploration in backbone models include the scaling law in model size, multilingual backbone, and model architectures.